\documentclass{article}

\usepackage[preprint]{neurips_data_2024}

\usepackage[utf8]{inputenc} 
\usepackage[T1]{fontenc}    
\usepackage{hyperref}       
\usepackage{url}            
\usepackage{booktabs}       
\usepackage{amsfonts}       
\usepackage{nicefrac}       
\usepackage{microtype}      
\usepackage{xcolor}         

\usepackage{graphicx}

\usepackage{natbib}
\setcitestyle{numbers,open={[},close={]},comma} 

\title{AICoderEval: Improving AI Domain Code Generation of Large Language Models}

\author{Yinghui Xia\\ AutoAgents.ai \\ \texttt{vix@autoagents.ai}
        \And
        Yuyan Chen \\ Fudan University \\ \texttt{chenyuyan21@m.fudan.edu.cn} \And
        Tianyu Shi \\ University of Toronto \\ \texttt{tianyu.s@outlook.com}
        \AND
        Jun Wang \\ East China Normal University \\ \texttt{wongjun@gmail.com} \And
        Jinsong Yang\thanks{Corresponsding author} \\ AutoAgents.ai \\ \texttt{edward.yang@autoagents.ai}
}

\begin{document}

\maketitle

\begin{abstract}
Automated code generation is a pivotal capability of large language models (LLMs). However, assessing this capability in real-world scenarios remains challenging. Previous methods focus more on low-level code generation, such as model loading, instead of generating high-level codes catering for real-world tasks, such as image-to-text, text classification, in various domains.
Therefore, we construct AICoderEval, a dataset focused on real-world tasks in various domains based on HuggingFace, PyTorch, and TensorFlow, along with comprehensive metrics for evaluation and enhancing LLMs' task-specific code generation capability. AICoderEval contains test cases and complete programs for automated evaluation of these tasks, covering domains such as natural language processing, computer vision, and multimodal learning.
To facilitate research in this area, we open-source the AICoderEval dataset at \url{https://huggingface.co/datasets/vixuowis/AICoderEval}.
After that, we propose CoderGen, an agent-based framework, to help LLMs generate codes related to real-world tasks on the constructed AICoderEval. 
Moreover, we train a more powerful task-specific code generation model, named AICoder, which is refined on llama-3 based on AICoderEval.
Our experiments demonstrate the effectiveness of CoderGen in improving LLMs' task-specific code generation capability (by 12.00\% on pass@1 for original model and 9.50\% on pass@1 for ReAct Agent).
AICoder also outperforms current code generation LLMs, indicating the great quality of the AICoderEval benchmark.
\end{abstract}

\begin{figure*}
  \centering
  \includegraphics[width=1\textwidth]{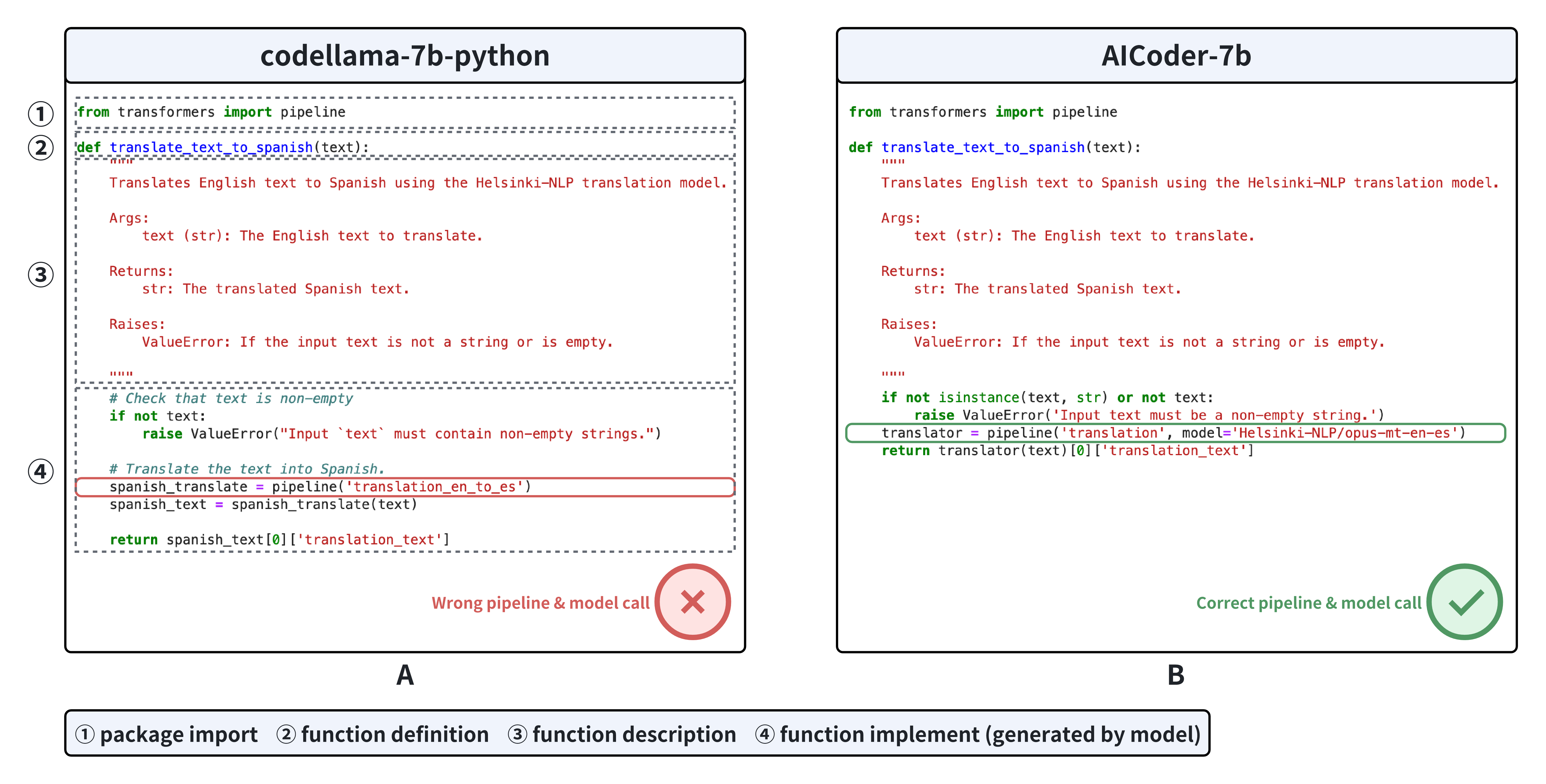}
  \caption{The AICoder generated by our CoderGen framework is capable of programming for domain-specific tasks and selecting the appropriate libraries for invocation. In part \textbf{A} depicts the output generated by codellama-7b-python, which incorrectly invoked a library using the pipeline method. In contrast, the part \textbf{B} presents the results produced by the AICoder, accurately selecting and calling the appropriate library to fulfill the requirements.}
  \label{fig:intro}
\end{figure*}

\section{Introduction}

Large language models attract attention for their general capabilities \cite{chowdhery2022palm, 10.5555/3495724.3495883, workshop2023bloom, touvron2023llama, du2022glm}, achieve high scores on evaluations such as HumanEval \cite{chen2021codex} and MBPP \cite{ni2023selfsampling}, which primarily focus on basic programming languages. However, their application capabilities in real software development, especially in the field of artificial intelligence using specific libraries (such as HuggingFace, PyTorch, TensorFlow, etc.), remain unclear. Although these libraries are very popular in AI development, how to evaluate and improve the code generation capabilities of large language models using these libraries is still a hard question.

Current researches explore how to leverage LLMs to use tool to call specific libraries. For instance, studies such as HuggingGPT \cite{shen2023hugginggpt} and Gorilla \cite{patil2023gorilla} try to generate single-line calls of APIs in specific domains. These studies show that even simple API calls require models to have a deep understanding and the ability to correctly use the libraries. However, these studies have not yet fully addressed how to automate the evaluation and enhancement of models' code generation capabilities in flexibly using specific libraries, especially when dealing with complex and diverse programming tasks.

To address this challenge, we construct the AICoderEval dataset, a benchmark for AI-oriented programming tasks to measure programming capabilities within this domain. AICoderEval covers a wide range of tasks across various AI domains, including natural language processing, computer vision, tabular data, audio and speech, multimodal learning, and reinforcement learning. The dataset is open-sourced and available at \url{https://huggingface.co/datasets/vixuowis/AICoderEval} to facilitate research in this area.

Then, we propose an agent-based framework called CoderGen, to generate task-specific codes. CoderGen simplifies the construction of datasets related to task-specific code on different libraries, enabling the automatic generation of training and testing samples. As illustrated in Figure \ref{fig:intro}, general code generation LLMs (e.g. codellama) may produce incorrect answers when it comes to pipeline and model API calls based on given function instructions. Our fine-tuned model demonstrates improved performance as it learns how to use the library for specific tasks. This approach allows for a more accurate assessment of a model's application capabilities in real software development and provides direction for further model improvements.

Our work includes three main contributions:

\begin{itemize}
    \item Benchmark Construction: We build the AICoderEval dataset, which focuses on AI tasks and includes code generation tasks related to AI libraries, along with test cases and complete programs for evaluating these tasks. These tasks cover a variety of library functions and usage patterns, ensuring that the model learns comprehensive knowledge about the libraries. We open-source the AICoderEval dataset at \url{https://huggingface.co/datasets/vixuowis/AICoderEval} to facilitate research in this area.
    \item Framework Design: We design and construct the CoderGen framework to generate high-quality training data. During the inference stage, we use an LLM-based agent to guide the generation of code that adheres to specific library usage standards, with continuous improvements in code quality. The agent interacts with the model multiple times to refine and optimize the code generation process, making it more consistent with library usage norms and best practices.
    \item Model Evaluation: We evaluate multiple large language models on AICoderEval, demonstrating their code generation capabilities in actual AI development tasks and the performance enhancements after training with our framework. This approach allows us to compare the performance of different models and identify their strengths and limitations in using specific libraries.
\end{itemize}

Through these contributions, CoderGen provides a more comprehensive and practical evaluation method for the code generation capabilities of large language models and points the way for further model improvements. We hope this framework will assist researchers and developers in better understanding and leveraging the potential of large language models in software development, particularly when programming with specific libraries.

\begin{table}
    \caption{Data Category Statistics}
    \centering
    \begin{tabular}{lll} 
        \hline
         \textbf{Category}&  \textbf{Cnt.} & \%\\ 
         \hline
         Natural Language Processing&  383 & 77.8\%\\ 
         Computer Vision&	50 & 10.2\%\\
        Tabular Data& 18 & 3.7\%\\
        Audio and Speech&	17 & 3.5\%\\
        Classification& 12 & 2.4\%\\
        Multimodal&	9 & 1.8\%\\
        Reinforcement Learning&	3 & 0.6\%\\
        \hline
        Total& 492 & 100\%\\
        \hline
         
    \end{tabular}
    \label{tab:category}
\end{table}

\section{Benchmark Construction}

\begin{figure*}[h]
  \centering
  \includegraphics[width=0.93\textwidth]{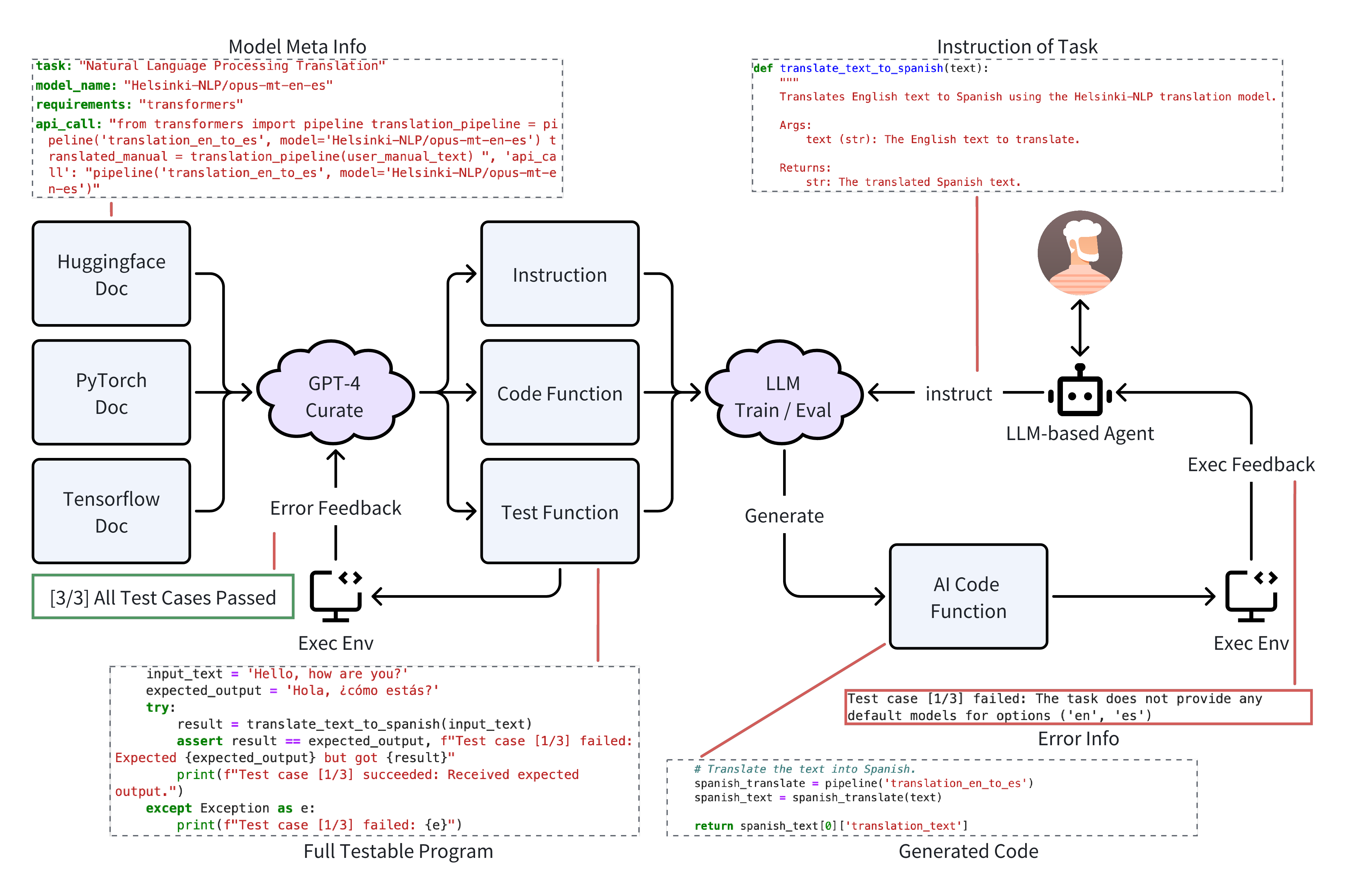}
  \caption{\textbf{CoderGen: A Domain-Specific Code Generation Architecture}. This architecture comprises two integral components. On the left side, \textbf{AICoderEval} data is produced by analyzing library documentation with provided document data (model meta-information). This data, which includes testable programs, is subsequently validated within an execution environment. We then utilize this data to train a LLM (\textbf{AICoder} in following paper). On the right side, an LLM-based agent is employed to direct the code generation process. Actual executable environments are utilized to push feedback to both the agent and the LLM, aiding in the refinement of the generated code.}
  \label{fig:main}
\end{figure*}

\subsection{Data Collection}

To construct the AICoderEval dataset, we leverage the power of GPT-4 \cite{openai2023gpt4} to process data collected from the web and format it into a structured form. Focusing on the domain of artificial intelligence, we select the Hugging Face Hub and PyTorch Hub as our target libraries. These libraries provide a unified API for model invocation, and their descriptions and documentation are readily available on the official websites. To reduce the complexity introduced by library descriptions, we directly employ data that has been automatically processed and filtered by GPT-4 as our input, which we then further process to create the desired dataset.

The web-scraped data, after filtering, contains the following information for each task: domain, model name, model description, example code, and performance metrics. This comprehensive set of information would enable human programmers to fully utilize it for development purposes. Consequently, we hypothesize that an intelligent agent should also be capable of learning to develop software based on these library specifications.

\subsection{Data Pre-processing and Curation}

To facilitate automated evaluation, we design our dataset structure inspired by the HumanEval \cite{chen2021codex} benchmark. Our primary focus is on generating Python code files using GPT-4, streamlining the process by concentrating on a single programming language. Each generated file is meticulously structured to encompass a comprehensive suite of components necessary for robust testing. These components include package installation instructions, package imports, main function definition, functionality description, input/output/error handling specifications, function implementation, testing functions, and test case invocations.

To ensure high-quality data generation, we provide GPT-4 with carefully designed in-context prompts and examples. These prompts are crafted to elicit the desired output format, which includes a problem description, an end-to-end solution utilizing specific library APIs, and a set of test cases. We also leverage GPT-4's function calling capability to generate the data in a piecewise manner, enhancing the stability and controllability of the output. This approach ensures that the generated code aligns well with the given prompts and passes the corresponding test cases. The exact prompts used for data generation can be found in Appendix \ref{sec:appendix:prompt}.

By consolidating the evaluation pipeline into a single code file for each task, we significantly simplify the testing process. All test cases for a given task can be executed by running the corresponding single file. Moreover, we prioritize diversity in the generated test cases, particularly in terms of difficulty levels. Through careful prompt engineering, we guide GPT-4 to generate three distinct test cases for each task: (1) a test for normal code execution, (2) a test for handling of edge cases and exceptional inputs, and (3) a test for verifying the correctness of the output under normal inputs.

After the initial data generation process, which produced approximately 9,000 code files, we proceed to filter and curate the dataset. We execute each code file in a sandboxed environment with GPU acceleration, retaining only the files that pass at least one test case. This filtering step reduces the dataset size to around 2,000 code files. To construct our final benchmark, we further select a subset of about 500 code files that pass all of their associated test cases. This rigorous filtering and curation process ensures the quality and reliability of the AICoderEval benchmark.

The AICoderEval benchmark is hosted on Hugging Face Datasets at \url{https://huggingface.co/datasets/vixuowis/AICoderEval}, a popular platform for sharing and discovering ML datasets. The dataset repository includes comprehensive documentation on the dataset structure, content, intended uses, and potential limitations. We also provide detailed instructions for accessing and using the dataset. The datasets are released under the permissive Apache License 2.0 to encourage broad adoption and facilitate future research. Moreover, we include structured metadata following the Hugging Face Datasets format to enhance discoverability and interoperability.

Table \ref{tab:category} presents the distribution of task categories within the AICoderEval benchmark. Natural Language Processing (NLP) tasks constitute the largest portion at 77.8\%, followed by Computer Vision (CV) tasks at 10.2\%. The remaining categories, including Tabular Data, Audio and Speech, Classification, Multimodal, and Reinforcement Learning, each account for less than 5\% of the total tasks. The NLP category encompasses a wide range of tasks such as text classification, text generation, and sentence similarity matching, while the CV category includes tasks like image classification, image segmentation, and image generation. This diverse set of tasks across various domains showcases the breadth and depth of the AICoderEval benchmark.

To ensure the quality and integrity of the AICoderEval benchmark, we will conduct both automated and manual auditing of the dataset. Automated checks will be performed to identify any sensitive or personally identifiable information, as well as to assess the diversity and balance of the tasks across different domains. Additionally, a subset of the generated code files will be manually reviewed by domain experts to verify their correctness, clarity, and adherence to best practices. Any issues or concerns identified during the auditing process will be promptly addressed before the official release of the benchmark.

\section{Methodology}

In this paper, we introduce CoderGen, an agent-based framework for generating codes on tasks in AICoderEval, as depicted in figure \ref{fig:main}. This framework can construct domain-specific tasks benchmark, for training and evaluation, and then fine-tunes a code generation model on the benchmark.

\subsection{Error Traceback and Analysis}

\begin{figure*}
  \centering
  \includegraphics[width=1\textwidth]{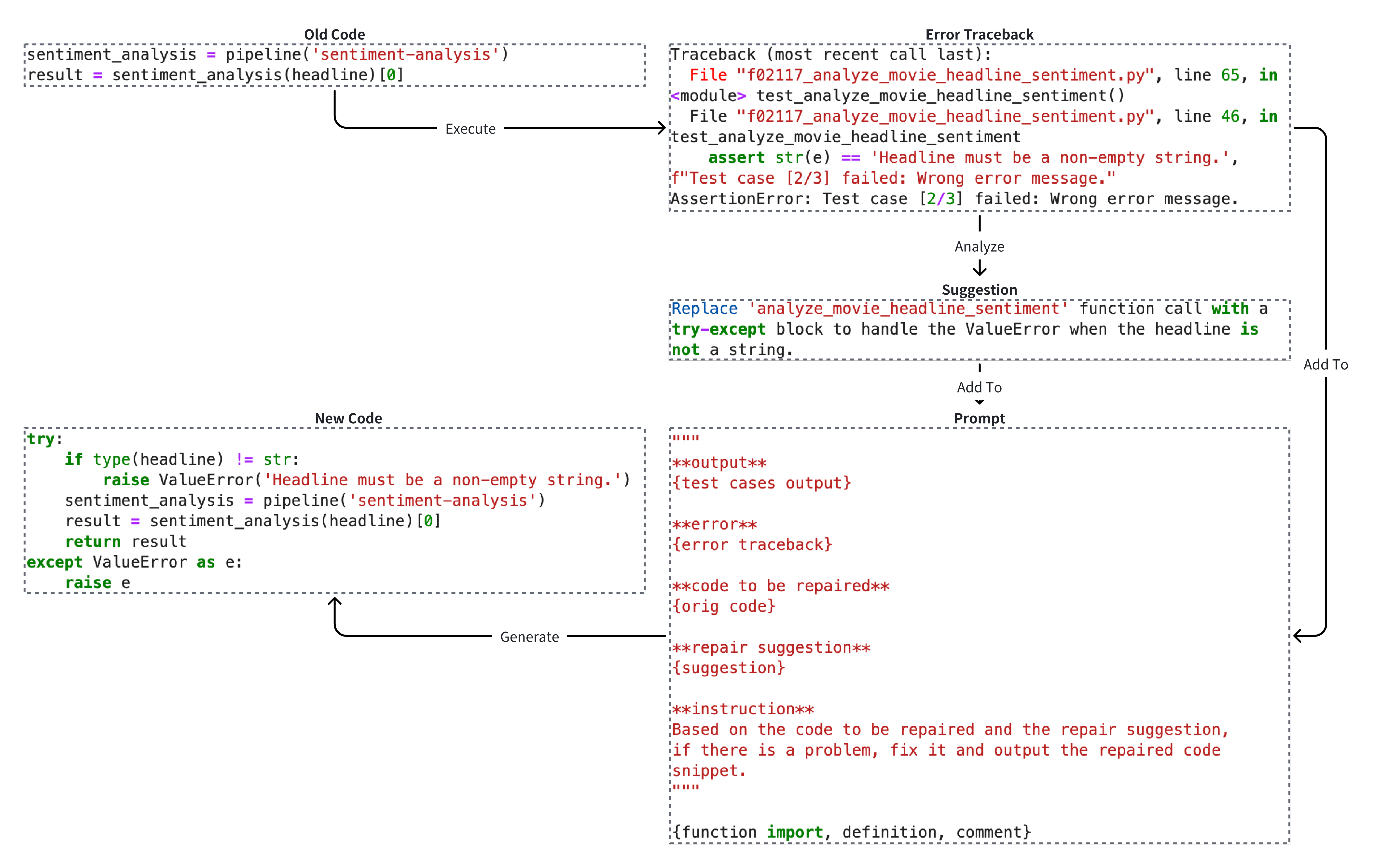}
  \caption{Error traceback analyze example}
  \label{fig:err-trace}
\end{figure*}

The CoderGen framework includes a robust error traceback and analysis mechanism to ensure that the generated code is not only syntactically correct but also functionally sound. Figure \ref{fig:err-trace} shows an example of error traceback and related prompt. After the initial code generation, the framework executes the code within a controlled environment to test its functionality. If the code fails to execute correctly, the system captures the error traceback, which provides a detailed record of the path through the code that led to the failure. This traceback is then analyzed by the framework to identify the specific point of failure, whether it be a syntax error, a logical error, or an issue with the code’s interaction with external libraries or APIs.

The error analysis component of CoderGen leverages the fine-tuned language model to interpret the error messages and suggest potential fixes. These suggestions are based on the model’s understanding of the code’s intended functionality and the context of the error within the broader codebase. The suggestions are then presented to the user, who can choose to implement them, or they can be automatically applied by the system for further testing. This iterative process of error detection, analysis, and correction continues until the code successfully executes all test cases and meets the specified requirements.

\subsection{Iterative Code Re-generation}

Once the errors have been identified and suggestions for improvement have been made, the CoderGen framework enters the code re-generation phase. Here, the framework uses the feedback from the error analysis to refine the code generation process. The erroneous code snippet, along with the suggestions and the original instruction, are fed back into the language model, which then generates a new version of the code snippet.

This new code snippet is then retested, and the process of error detection, analysis, and correction is repeated. This iterative cycle ensures that the generated code not only resolves the immediate issues but also improves in quality and robustness with each iteration. The framework’s ability to learn from its mistakes and adapt its code generation strategy based on real-time feedback is a key feature that sets CoderGen apart from traditional code generation systems.

By incorporating these iterative feedback loops, CoderGen aims to produce code that is not only correct but also efficient and maintainable, reflecting the best practices and idioms of the target domain. This approach has the potential to significantly reduce the time and effort required for developers to produce high-quality code, particularly in complex and specialized domains.

\section{ Experiment }

\subsection{Experimental Setup }

We perform inference on a single NVIDIA GeForce RTX 4090 GPU using various models, including the first stage of completion and the second stage of bug fixing. For the parameter settings, we set the top-p value to 0.9, the temperature parameter to 0.6, and the maximum token count to 2048.

To further validate that the model's performance can be enhanced through specialized datasets, we host the training data on Hugging Face. We employ the LoRA (Low-Rank Adaptation) technique, which operates on the principle of adding trainable low-rank matrices to the original model's weights, allowing for efficient fine-tuning with minimal additional parameters. We utilize the PEFT (Parameter-Efficient Fine-Tuning) framework to fine-tune the model. Due to LoRA's rapid convergence properties, we conduct the fine-tuning process using a single NVIDIA GeForce RTX 4090 GPU. We use LoRA parameters with a rank of 8 and an alpha value of 32. The learning rate is set to 1e-4, with a batch size of 4, and we train the model for 3 epochs, utilizing gradient accumulation steps to optimize computational efficiency and resource utilization.

By leveraging specialized datasets and employing parameter-efficient techniques like LoRA, we aim to further improve the model's performance and capabilities. The combination of high-performance hardware, carefully selected hyperparameters, and efficient fine-tuning frameworks enables us to push the boundaries of what our models can achieve in tasks such as code completion and bug fixing.


\subsection{Main Results}



\begin{table*}[h]
  \caption{Experiment on \textbf{AICoderEval} dataset}
  \centering
  \begin{tabular}{llllllllllllllll}
    \hline
    Models & \multicolumn{2}{c}{Original} & \multicolumn{2}{c}{w/ ReAct Agent} & \multicolumn{2}{c}{Relative Increase}\\
    & \small{SR@All} & \small{SR@Any} & \small{SR@All} & \small{SR@Any} & \small{SR@All ↑\%} & \small{SR@Any ↑\%} \\
    \hline

    GPT-3.5-turbo-1106 & 9.16 & 46.84 & 13.03 & 60.63 & 42.25 & 29.44  \\

    llama-2-7b & 1.23 & 26.02 & 1.83 & 33.41 & 48.78 & 28.40 \\
    llama-2-13b & 2.76 & 42.04 & 3.98 & 51.24 & 44.20 & 21.88 \\
    llama-2-70b & 6.32 & 65.89 & 8.16 & 78.68 & 29.11 & 19.41 \\

    codellama-7b-python & 19.58 & 66.95 & 23.86 & 78.18 & 21.86 & 16.77  \\
    codellama-13b-python & 20.46 & 67.22 & 23.88 & 75.67 & 16.72 & 12.57 \\
    codellama-34b-python & 23.68 & 70.19 & 25.78 & 77.33 & 8.87 & 10.17\\

    llama-3-8b-instruct & 30.49 & 85.80 & 32.11 & 86.82 & 2.96 & 1.19
    
    \\

    \hline
    llama-3-8b-instruct w/ sft & \textbf{34.15} & \textbf{86.18} & \textbf{35.16} & \textbf{86.99} & 2.96 & 0.94 \\
    ↑ (v.s. w/o sft) & 3.66 & 0.38 & 3.05 & 0.17 & - & - \\
    ↑\% & 12.00 & 0.44 & 9.50 & 0.20 & - & - \\

    \hline
  \end{tabular}
  \label{tab:table-1}
\end{table*}

In this study, we utilized the \textbf{AICoderEval} dataset to test multiple popular API and open-source LLM models, particularly those equipped with code generation capabilities. The models tested included gpt-3.5-turbo-1106 supported by OpenAI, as well as Llama 2 7b / 13b / 70b \cite{touvron2023llama}, llama 3 8b, and Codellama 7b / 13b / 34b \cite{rozière2024code} models developed by Meta. Furthermore, we fine-tuned \textbf{AICoder} based on the llama-3-8b-instruct model. Table \ref{tab:table-1} presents a comparison of these models' performance in their original versions and after the introduction of an error repair agent, where \textbf{SR@All} represents the success rate of all tests passed for a single program, and \textbf{SR@Any} represents the success rate of any test case passed for a single program.

In this study, we conducted an experiment using the AICoderEval dataset to evaluate several prominent API and open-source Large Language Models (LLMs) with code generation capabilities. The introduction of an error repair agent (ReAct Agent) and supervised fine-tuning (sft) significantly enhanced the performance metrics of all tested models, particularly SR@All and SR@Any. On average, the task-specific code generation capabilities of LLMs improved by approximately 28.20\% for SR@All and 18.60\% for SR@Any after the ReAct Agent was introduced. For example, GPT-3.5-turbo-1106 saw an increase in SR@All from 9.16\% to 13.03\% and in SR@Any from 46.84\% to 60.63\%.

We also noted a direct correlation between model scale and the extent of performance improvement, with larger models in the Llama 2 series showing greater enhancements in SR@All and SR@Any, such as llama-2-70b outperforming llama-2-7b. Supervised fine-tuning further boosted the performance of the llama-3-8b-instruct model, with a 2.96\% increase in SR@All and a 0.94\% increase in SR@Any.

Moreover, domain-specific fine-tuning significantly elevated the performance of the original network, as evidenced by AICoder-7b surpassing all tested baseline models in both SR@All and SR@Any metrics, achieving state-of-the-art results.

Overall, this research underscores the effectiveness of error repair agents and supervised fine-tuning in advancing the code generation capabilities of LLMs. By carefully selecting fine-tuning strategies and taking into account model scale, substantial improvements in model performance on specific tasks can be achieved. These insights are instrumental for guiding future optimizations in code generation and beyond.

\begin{table}[h]
  \caption{Experiment on \textbf{AICoderEval} dataset. \textbf{CL} is for average code lines, and \textbf{CT} is for average code tokens}
  \centering
  \begin{tabular}{llll}
    \hline
    Models & Code Lines (CL) & Code Tokens (CT) & Rank \\
    \hline

    GPT-3.5-turbo-1106 & 8.6 & 62.9 & 1\\

    llama-2-7b & 16.2 & 112.9 & 5\\
    llama-2-13b & 18.5 & 116.3 & 7\\
    llama-2-70b & 13.1 & 107.8 & 4\\
    
    codellama-7b-python & 21.5 & 128.3 & 9\\
    codellama-13b-python & 18.9 & 116.3 & 8\\
    codellama-34b-python &  18.4 & 114.4 & 6\\

    llama-3-8b-instruct & 11.02 & 96.97 & 3\\
    
    \hline
    llama-3-8b-instruct w/sft & 9.32 & 87.71 & 2\\
    

    \hline
  \end{tabular}
  \label{tab:table-2}
\end{table}

Table \ref{tab:table-2} shows the number of code lines (\textbf{CL}) and code tokens (\textbf{CT}) generated by different models. We can identify a pattern where shorter code generated by the models typically implies stronger problem-solving abilities and more concise solutions. For instance, codellama-34b-python had lower CL and CT than codellama-7b-python, which aligns with its relative performance in SR@All and SR@Any, while AICoder outperformed with significantly shorter generated code lines compared to other models.



In summary, the introduction of the error repair agent has significantly improved the overall performance of the models, whether it is the success rate of all tests passed for a single program (SR@All) or any test case passed (SR@Any). The increase in model scale has a positive impact on performance improvement, especially in the Llama 2 series where larger model scales result in more pronounced performance gains. The fine-tuning strategy has also demonstrated its effectiveness, particularly for the AICoder model, which achieved state-of-the-art performance in all tested baselines after fine-tuning. The performance of the models varies significantly across different task categories, indicating the necessity for domain-specific optimization and improvement.

\section { Related Work }

\subsection{Code Generation}

Utilizing language models for code generation is a challenging task \cite{doi:10.1126/science.abq1158, 10.1145/3520312.3534862, 10.1145/3510003.3510203}. Researchers propose various methods to enhance the capabilities of language models in programming tasks, including task decomposition \cite{kim2023language, yao2023react}, self-debug \cite{chen2024teaching}, and code generation models. These efforts primarily focus on the generation of general code, with less attention given to the capabilities of domain-specific code. In real-world scenarios, however, we often use libraries to create new tools and implement more complex functionalities through longer chains of function calls. Therefore, our research aims to enable programs to automatically solve tasks using domain-specific libraries and to verify the results automatically, thereby expanding the capabilities of code generation.

\subsection{Tool Usage}

Large language models can leverage tools to enhance their capabilities, such as Toolformer \cite{Schick2023ToolformerLM} and GPT-4 \cite{openai2023gpt4} making API calls more feasible. Traditional tools include web browsing, calculators, code interpreters, etc., with these efforts aiming to invoke general capabilities. HuggingGPT \cite{shen2023hugginggpt} and Gorilla \cite{patil2023gorilla}, on the other hand, focus on domain-specific API calls. Our research aims to explore the programming capabilities of specific domain libraries, thereby expanding the scope of program usability.

\subsection{Agent}

An agent is generally represented as an entity with the capability to interact with the environment and take actions, either based on feedback from the environment or driven by intrinsic motivations. It exhibits greater adaptability and versatility in its capabilities and execution outcomes compared to ordinary programs. LLM-based Agents have recently been widely discussed \cite{xi2023rise, wang2023interactive, Park2023GenerativeAgents}; they expand their capabilities through the use of tools, and planning ability is also one of the most important capabilities of LLM-based Agents. In the field of code generation, previous work has focused more on one-time code generation, such as CodeGen \cite{nijkamp2022codegen}, CodeX \cite{chen2021codex}. However, in real-world scenarios, we approach the expected results incrementally through feedback from the actual environment, such as execution information and error messages. In this paper, our research aims to enable Agents to analyze error messages, allowing the program to execute correctly.

\section{Conclusions and Future Work}

This paper introduces CoderGen, an automatic learning and evaluation framework designed to improve the assessment of code generation capabilities, especially when dealing with libraries commonly used in real software development. CoderGen automatically constructs an evaluation dataset, AICoderEval, for libraries related to artificial intelligence, and trains a domain-optimized code generation model based on this dataset. Furthermore, the AICoder model is fine-tuned on the codellama dataset and evaluated on the AICoderEval dataset, demonstrating its superiority over other code generation models. Our work represents a significant advancement in evaluating and enhancing code generation capabilities in real software development by focusing on the understanding and application of libraries commonly used in actual software development processes. In future work, we plan to optimize the CoderGen framework to support a wider range of libraries and software development scenarios, validate its generality and effectiveness with diverse datasets and tasks, and integrate it with the latest code generation technologies to further enhance model performance and practicality.

\section*{Limitation}

The CoderGen framework makes great strides in evaluating code generation skills, but it currently has some limitations. First, it mainly uses a dataset on AI specific tasks, so it needs more testing to see if it works well for other types of software development. Second, even though we improve the AICoder model with the codellama dataset, it could still be better, and we need to keep working on it. Lastly, our testing method is simple and needs to be more robust for testing, possibly by using Docker and cloud platforms to make it easier for others to repeat our tests and build on our work.

{
\small
\bibliography{custom}
}

\newpage

\newpage
\appendix

\begin{table*}
\section{Appendix}
\subsection{Prompt Details}

\caption{Prompt details of GPT-4 dataset generation. Combining all parts from table into a complete prompt enables GPT-4 to convert domain documents into an executable code dataset.}

\label{sec:appendix:prompt}

    \centering
    \begin{tabular}{ll}
        \hline
        \small{\textbf{Task Prompt}}& 

    \begin{minipage}[l]{1.5\columnwidth}
    \begin{verbatim}

1. Please design a requirement that can be described in one sent-
ence.
2. Based on the above description, generate code to implement the
requirement.
3. Function comments should follow the Google Python Style Guide, 
including args, returns, and raises.
4. Write corresponding test functions based on the generated code.
5. The test cases should be three examples of different difficulty
levels, e.g., the first one verifies that the function executes 
normally, the second verifies that incorrect inputs are handled
properly, and thethird verifies that the function returns the cor-
rect value.
6. For testing purposes, read image and audio files, download 
them from online resources to the local machine, or obtain them 
from datasets; do not provide fake or non-existent file addresses.
    \end{verbatim}
    \end{minipage}
        \\
        \hline

        \small{\textbf{Import example}}& 

    \begin{minipage}[l]{1.5\columnwidth}
    \begin{verbatim}
    
import subprocess
requirements = ["package1", "package2"]
for package in requirements:
    subprocess.run(['pip', 'install', '-U', package])
    \end{verbatim}
    \end{minipage}
        \\
        \hline

        \small{\textbf{Test prompt}}& 

    \begin{minipage}[l]{1.5\columnwidth}
    \begin{verbatim}

1. The function starts by printing "Testing started."
2. For images or audio,  load a dataset or download data from on-
line resources.
3. The test case starts by printing "Testing case [x/x] started", 
prints "succeeded" on success, and "failed" on failure.
4. The function ends by printing "Testing finished."
    \end{verbatim}
    \end{minipage}
        \\
        \hline

        \small{\textbf{Test example}}& 

    \begin{minipage}[l]{1.5\columnwidth}
    \begin{verbatim}

def test_...():
    print("Test started.")
    dataset = load_dataset("...")
    sample_data = dataset[0]  # Extract a sample from the dataset
    
    # Test case 1:...
    print("Test case [1/3] started.")
    try:
        assert assert 1, f"Test case [1/3] failed: ..."
        print(f"Test case [1/3] succeeded: ...")
    except Exception as e::
        print(f"Test case [1/3] failed: ...\nerror:", e)

    # Test case 2:...
    
    # Test case 3:...

# Run the test function
test_...()
    \end{verbatim}
    \end{minipage}
        \\
        \hline

    \end{tabular}
    \label{tab:data-prompt}
\end{table*}

\newpage

\end{document}